\documentclass{article}
\usepackage{hyperref, spconf,amsmath,graphicx,float, subfigure}
\usepackage{algorithm, algorithmicx}  
\usepackage{algpseudocode}  
\usepackage{multirow}
\usepackage{amssymb, bm}

\newcommand{\Tref}[1]{Table~\ref{#1}}

\newcommand{\Fref}[1]{Fig.~\ref{#1}}

\newcommand{\Sref}[1]{Section~\ref{#1}}

\graphicspath{{figures/}}

\title{Training-Free Location-Aware Text-to-Image Synthesis}
\twoauthors
 {Jiafeng Mao}
	{The University of Tokyo\\
        Dept. of Information and Communication Eng.\\
	mao@hal.t.u-tokyo.ac.jp}
  {Xueting Wang}
	{CyberAgent, Inc.\\
	AI Lab \\
	wang\_xueting@cyberagent.co.jp}
\begin{document}
%
\maketitle
\begin{abstract}
\vspace{-5pt}
Current large-scale generative models have impressive efficiency in generating high-quality images based on text prompts. However, they lack the ability to precisely control the size and position of objects in the generated image. In this study, we analyze the generative mechanism of the stable diffusion model and propose a new interactive generation paradigm that allows users to specify the position of generated objects without additional training. Moreover, we propose an object detection-based evaluation metric to assess the control capability of location aware generation task. Our experimental results show that our method outperforms state-of-the-art methods on both control capacity and image quality.
\end{abstract}

\vspace{-5pt}
\section{Introduction}
\label{sec:intro}
\vspace{-5pt}
Text-to-image synthesis is a rapidly evolving field aimed at generating natural and realistic images based on textual prompts. The emergence of generative diffusion models~\cite{ho2020denoising} and related research~\cite{dhariwal2021diffusion, song2020denoising, rombach2022high, ramesh2022hierarchical, nichol2021improved, nichol2022glide} has significantly improved the quality of generated images. Recent research has focused on enhancing the quality of generated images and diverse fine-grained control on the editability of image content by diffusion models~\cite{liu2022compositional, kawar2022imagic, feng2022training, hertz2022prompt, park2021benchmark, balaji2022ediffi}, for example, Compositional diffusion~\cite{liu2022compositional} allows user to give multiple concepts and generate images with feature combinations, and Imagic~\cite{kawar2022imagic} allows user to edit semantic attributes on input images by specified test. However, seldom attention has been paid to fine-grained control of object location and size within the generated content. 

\begin{figure}[t]
    \centering
    \includegraphics[width=0.46\textwidth]{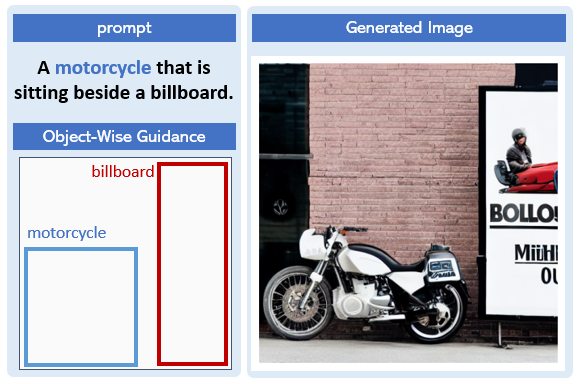}
    \caption{Concept of our task setting. Users can provide both prompt and exact region guidance of objects. Then the generation model aims at generating images where each object is positioned in its specified location.}
    \label{fig:setting}
    \vspace{-10pt}
\end{figure}

This work aims to make generated images more precisely aligned with user expectations. Whereas typical text-guided diffusion models accept textual prompts, we propose a new task setting that also accepts the specific location and size of objects described in the prompt and generates images where each object is positioned in the location specified by users as they want, as shown in \Fref{fig:setting}. This setting contribute to the accurate user-aligned generation and more stable generation performance of multiple objects.

\begin{figure*}[t]
    \centering
    \includegraphics[width=1\textwidth]{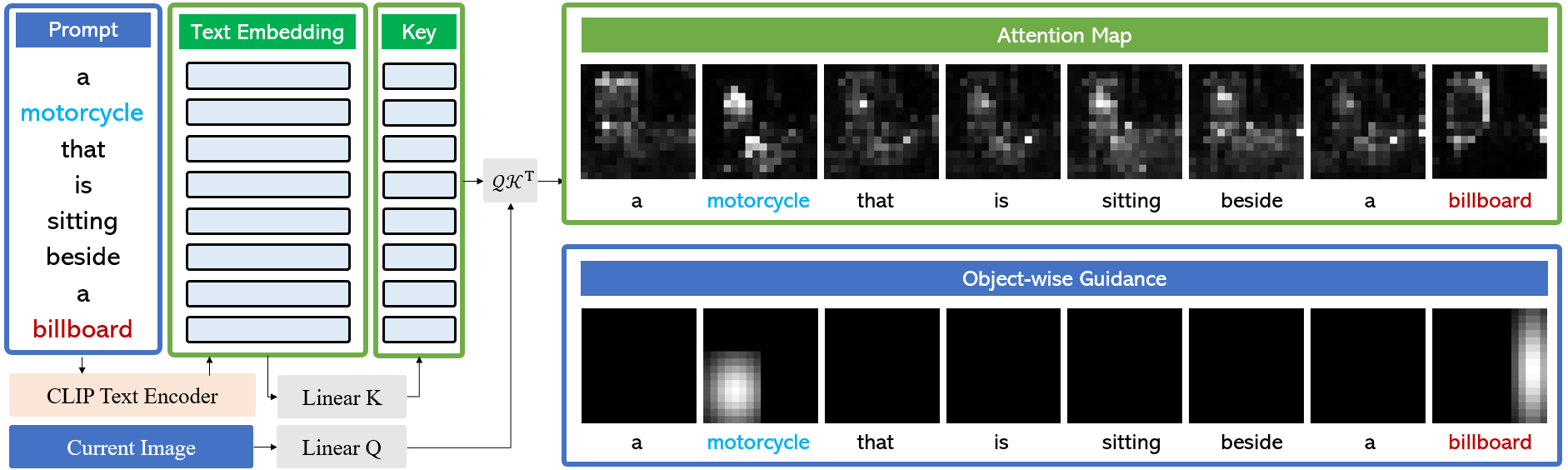}
    \vspace{-20pt}
    \caption{Illustration of our proposal. The text prompt is processed into a text embedding using CLIP, which is then used to compute the attention map. Each entry of the attention map corresponds to a word in the text prompt. The object-wise guidance is processed into a soft mask~(\Sref{sec:soft}) with out-of-region suppression~(\Sref{sec:out}) and added to the attention map to increase the likelihood of the corresponding concept features appearing in specific locations.}
    \label{fig:mask}
\end{figure*}

Existing research~\cite{hertz2022prompt} has observed that the cross-attention layer~\cite{vaswani2017attention} of the U-Net~\cite{ronneberger2015u} in diffusion models controls the layout and structure of generated images. Therefore, we propose controlling the position of individual objects in generated images by manipulating the values of cross-attention layers without additional training. Currently, no evaluation metric is designed to measure the control efficiency for the object-wise location-guided generation. We also propose a comprehensive evaluation method based on object detectors to calculate the consistency of object information between generated images and guidance information. We conducted experiments using the MS COCO dataset~\cite{lin2014microsoft} to show the effectiveness of our proposal on both object-wise control efficiency and image quality. We summarize the contributions of this paper as follows:
\vspace{-5pt}
\begin{itemize}
    \setlength\itemsep{0pt}
    \item We propose an intuitive yet effective method to control the region of objects on generated images, allowing users to specify the precise position and size of the objects they wish to generate.
    \vspace{-5pt}
    \item We propose an evaluation method to estimate the object-wise consistency between the generated images and the guidance information for location-aware text-to-image generation tasks.
    \vspace{-10pt}
\end{itemize}

\section{Related Works}
\label{sec:related works}

Some methods have been proposed to improve the controllability of diffusion models by making it easier for users to specify the content of generated images. Imagic~\cite{kawar2022imagic} focuses on attribute editing, designed to generate edited images according to the specified text. Compositional diffusion~\cite{liu2022compositional} takes multiple text conditions and outputs an image containing features of all text inputs. Unlike these works, our focus is on generating images where objects are in the specific locations users want them to be. Paint-with-Word~\cite{balaji2022ediffi} shows that the location of generated objects can be manipulated by changing the attention maps. However, according to our experiments, its mask can harm the generation quality. Inspired by it, we propose a finely designed method of manipulating attention maps to control the region of generated objects.

\section{Proposed Method}
\label{sec:proposal}
\subsection{Problem Setting}
\label{sec:problem setting}
In our task, we aim to generate images not only according to a given text prompt $\mathcal{P}$, which is an image-wise caption, but also complying with the object-wise guidance $\{\mathcal{B},\mathcal{C}\}$, where $\mathcal{B}$ denotes a list of bounding boxes indicating the desired spatial location of each object and $\mathcal{C}$ is a list of the object category mentioned in prompt $\mathcal{P}$.

\subsection{Stable Diffusion}

\label{sec:stable}
Our approach and experiments utilize the state-of-the-art text-to-image model, Stable Diffusion~\cite{rombach2022high} as the baseline model. This model achieves high performance by projecting noisy images into a latent space, using a diffusion model to remove the noise. The denoised images in the latent space are then decoded as the final output. To accomplish this, Stable Diffusion employs a modified U-Net for noise estimation and a frozen CLIP~\cite{radford2021learning} text encoder to convert text inputs into embedding sequences. The relationship between the image and text is established through multiple cross-attention layers in the down-sampling and up-sampling stages. During a generation of an image, the CLIP encoder encodes provided prompt $\mathcal{P}$ as an embedding sequence, which is subsequently processed into keys and values $K, V \in R^{(n,l)}$ in each cross-attention block by linear layers. The latent $\mathcal{X}$ of the current image is also projected into a query $Q \in R^{(n, [w, h])}$ in each cross-attention block, and the attention maps $M \in R^{([w, h], l)}$ are calculated by $Q$ and $K$ as follows,
\begin{equation}
    M = \text{Softmax}(\frac{QK^T}{ \sqrt{d} }).
\end{equation}
The attention map $M$ controls the spatial distribution of values $V$, which contains rich semantic information about the image content specified by the prompt $\mathcal{P}$. It is indicated that each attention map entry highly corresponds to a single word in the prompt~\cite{hertz2022prompt}. In an attention map, each entry corresponds to a word in the text prompt. When the attention map for a particular word has a high value in a specific region, this region will likely present the features of the concept associated with that word.

\subsection{Location-Aware Text-to-Image Synthesis}
\label{sec:spatial control}

A straightforward way to manipulate the location of generated objects is by increasing the values of specified concepts in the specified area on the attention maps. Given a prompt $\mathcal{P}$ and a object-wise guidance $\{b, c\} \in \{\mathcal{B}, \mathcal{C}\}$, since the $c \in \mathcal{C}$ is a word mentioned in $\mathcal{P}$, there is a particular entry of attention map $M_c$ corresponding to $c$. We proposed to increase the value in area $b$ on the $M_c$ by a gaussian-based soft mask $\mathcal{G}$ so that image tokens in the region $b$ are encouraged to attend more to the text token of $c$. As a result, the semantic concept corresponding to $c$ is more likely to appear in the area $b$. Specifically, the calculation of attention maps is modified as follows,
 \begin{equation}
     M=\text{Softmax}(\frac{QK^T + wS}{ \sqrt{d} })
 \end{equation}
where $S$ is our proposed soft mask and has the same shape as $QK^T$. Values in $S$ that are specified by $\{\mathcal{B}, \mathcal{C}\}$ are calculated by a gaussian mask $\mathcal{G}$, while others are set to $-\infty$, as shown in \Fref{fig:mask}. Specifically, for each item $\{b, c\}$ in object-wise guidance $\{\mathcal{B},\mathcal{C}\}$, the mask $S_c$ is calculated as follows,
\begin{equation}
\label{eq:A}
    S_c~(x, y)= 
    \begin{cases}
        \mathcal{G}_c(x,y) & \text{If $(x,y)$ inside $b$} \\
        -\infty & \text{Otherwise}
    \end{cases}
\end{equation}
$w$ is the weight of the mask, which is calculated as follows,
\begin{equation}
\label{eq:w}
    w = w^{\prime} \cdot f(T) \cdot \text{max}(Q \cdot K^T),
\end{equation}
where $f(T)$ is a linear function of the denoising step $T$ that becomes smaller as the noise level decreases. $w^{\prime}$ is a user-specified scalar.
We then introduce the soft mask $S_c$ in detail as followings.

\subsubsection{Gaussian-based Mask}
\label{sec:soft}
\begin{figure}[t]
    \centering
    \includegraphics[width=0.48\textwidth]{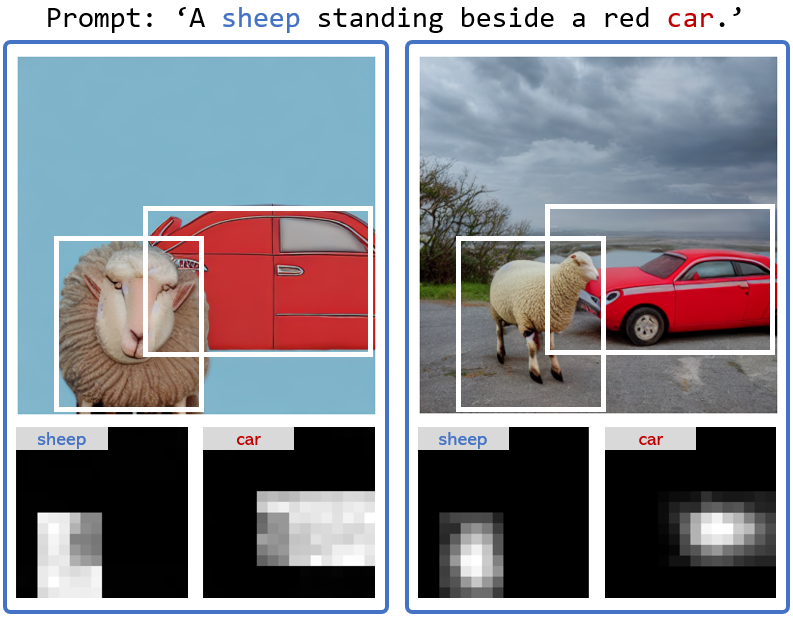}
    \vspace{-20pt}
    \caption{Generated images under object-wise guidance. The left image is guided by Paint-with-words~\cite{balaji2022ediffi}, and the right image is guided by our soft mask.}
    \label{fig:compare}
\end{figure}

By adding values to the specified area, Paint-with-words~\cite{balaji2022ediffi} makes the specified words dominate the area and prevents attention maps for other words from working in that area. This approach impacts the internal distribution of attention maps and therefore affects the quality of generated images, particularly when the size of the specified object is large. Our proposal can effectively alleviate the negative impact caused by additional masks while maintaining the control efficiency, as shown in \Fref{fig:compare}. We model our soft mask based on a Gaussian probability density function. Similar to a Gaussian distribution, our soft mask will have high values in the central region and low values at the edges. Thus, our mask can retain the intensity of the guided generation using the high values in the central region while reducing the impact on surrounding regions. Specifically, for each item $\{b, c\}$ in object-wise guidance $\{\mathcal{B},\mathcal{C}\}$, where b is denoted as $\{x_{min}, y_{min}, x_{max}, y_{max}\}$, our soft mask $\mathcal{G}_c(x,y)$ for concept $c$ is calculated as follows,

\begin{equation}
\label{eq:Ac}
    \mathcal{G}_c(x,y)= \frac{1}{2\pi}e^{-\frac{1}{2}(D(x)^2+D(y)^2)}
    .
\end{equation}
where
\begin{equation}
\label{eq:dx}
    D(x)=\frac{2x-(x_{max}+x_{min})}{(x_{max}-x_{min})}s,\\
\end{equation}
and 
\begin{equation}
\label{eq:dy}
    D(y)=\frac{2y-(y_{max}+y_{min})}{(y_{max}-y_{min})}s,
\end{equation}
where $s$ is a hyper-parameter to control the softness of the mask and is set to 2 experimentally. Larger $s$ will lead to smaller values on the edge of the mask.

\vspace{-10pt}
\subsubsection{Out-of-region Suppression}

\label{sec:out}
Using a specific concept's mask to enhance the value of specific areas in the attention map can increase the likelihood of this concept appearing in this area. However, when the attention map already has high values outside the specified area, only enhancing the values of the specific area cannot prevent the objects from being generated outside the area. Thus, we set the values of the mask outside the area to $-\infty$ to suppress the generation of out-of-region objects. After softmax, the values outside the specified area are calculated as $0$ so that the generated objects are restricted to the specified area. 

\begin{table}[t]
  \begin{center}
    \setlength{\tabcolsep}{3.7mm}{
  \begin{tabular}{l|c|cc|c}
    \hline
     methods      &  $w^{\prime}$  & IoU$\uparrow$  & $R_{\text{suc}}\%\uparrow$ & FID$\downarrow$    \\
    \hline
    \hline
    Stable~\cite{rombach2022high}  & - & $0.19$ & $12.14$ & $20.78$ \\
    \hline
   \multirow{4}*{Paint~\cite{balaji2022ediffi}}   & $0.05$ & $0.22$  & $14.86$ & $21.45$ \\
    ~      & $0.1$ & $0.24$  & $17.56$  & $22.72$ \\
    ~  & $0.15$ & $0.22$ & $15.38$  & $27.07$ \\
    ~  & $0.2$  & $0.16$  & $11.34$ & $47.73$ \\
    \hline
    \hline
    \multirow{2}*{$\text{Ours}$}  & $0.15$ & \underline{$0.25$} & \underline{$19.24$} & \bm{$20.40$} \\
   ~   &$0.2$ & \bm{$0.27$}  &  \bm{$22.31$} & \underline{$21.35$} \\
    \hline
  
  \end{tabular}}
\end{center}
\vspace{-15pt}
\caption{Results on selected MS COCO dataset. Our proposal significantly outperformed the state-of-the-art on control efficiency and image quality. Columns of IoU and $R_\text{suc}$ indicate the average recorded IoU and the ratio of successfully generated objects to all objects, respectively.}
\label{tab:results}
\vspace{-10pt}
\end{table}

\begin{table*}[t]
  \begin{center}
    \setlength{\tabcolsep}{4.4mm}{
  \begin{tabular}{l|c|cccc|cccc}
    \hline
     methods     & subset  & IoU  &  $\text{IoU}_s$  & $\text{IoU}_m$ & $\text{IoU}_l$ & $R_{\text{suc}}\%$ & $R^s_{\text{suc}}\%$ &$R^m_{\text{suc}}\%$ &$R^l_{\text{suc}}\%$     \\
    \hline
    \hline
   \multirow{4}*{Stable~\cite{rombach2022high}}   & $near$ & $0.29$ & $0.07$ & $0.27$ & $0.41$ & $24.20$ & $0.86$ & $12.26$ & $45.09$ \\
   ~   & $mid$ & $0.19$ & $0.06$ & $0.23$ & $0.35$ & $10.23$ & $0.50$ & $8.84$ & $30.71$\\
   ~   & $far$ & $0.09$ & $0.04$ & $0.15$ & $0.28$ & $2.32$ & $0.39$ & $3.99 $ & $22.08$\\
   ~   & $All$   & $0.19$ & $0.05$ & $0.21$ & $0.38$ & $12.14$ & $0.50$ & $8.11$ & $39.41$ \\
    \hline
    \multirow{4}*{Paint~\cite{balaji2022ediffi}}  & $near$ &  $0.34$ & $0.09$ & $0.33$ &  $0.48$ & $31.10$ &  $1.01$ & $18.39$ &  $56.01$ \\
   ~                      & $mid$  &  $0.24$ & $0.09$ & $0.28$ &  $0.43$ & $16.17$ &  $1.42$ & $14.15$ &  $47.05$\\
    ~                     & $far$  &  $0.13$ & $0.07$ & $0.22$ &  $0.38$ & $5.65$ &  $1.10$ & $10.47$ &  $37.66$\\
    ~  &  $All$  &  $0.24$ & $0.08$ & $0.27$ & $0.46$ & $17.56$ & $1.16$ & $14.05$ & $52.34$ \\
    \hline
    \hline
    \multirow{4}*{ours}  & $near$ & \bm{$0.37$} & \bm{$0.12$} & \bm{$0.37$} & \bm{$0.50$} & \bm{$37.03$} & \bm{$3.74$} & \bm{$27.82$} & \bm{$60.92$}\\
   ~                     & $mid$ & \bm{$0.27$} & \bm{$0.10$} & \bm{$0.32$} & \bm{$0.45$} & \bm{$20.58$} & \bm{$2.59$} & \bm{$22.48$} & \bm{$49.32$}\\
    ~                    & $far$ & \bm{$0.17$} & \bm{$0.09$} & \bm{$0.26$} & \bm{$0.40$} & \bm{$9.75$} & \bm{$2.97$} & \bm{$17.94$} & \bm{$41.56$}\\
    ~                    & $All$ & \bm{$0.27$} & \bm{$0.10$}  & \bm{$0.31$}  & \bm{ $0.48$ } &  \bm{$22.31$} & \bm{$2.95$}  &  \bm{$22.38$} & \bm{$56.24$} \\
    \hline
  \end{tabular}}
\end{center}
\vspace{-15pt}
\caption{Control efficiency of guidance on different positions. Our method achieved the best performance on all subsets.}
\vspace{-15pt}
\label{tab:distance}
\end{table*}

\vspace{-10pt}
\section{Experiments}
\label{sec:typestyle}
\subsection{Datasets}
We use the MS COCO dataset, which contains image caption and object annotation, to validate our proposed method. We selected samples from the MS COCO dataset that meet the requirements of our task, i.e., samples whose object-wise annotations are all mentioned in the image caption (prompt). Since most of the image captions involve only one or two significant objects in the image, we randomly selected 6000 images, where 3000 contain one object and the other 3000 contain two objects. All images are resized to $512^2$.

\vspace{-10pt}
\subsection{Evaluation Metrics}

\noindent\textbf{Object-wise Consistancy} We propose an object detector-based evaluation method to quantify the effectiveness in controlling the positions of generated objects. We generate images using image captions as prompt and object-wise bounding boxes from the dataset as object-wise guidance. Then, we use a YOLOR~\cite{wang2021you} to perform object detection on each generated image and obtain the class labels and bounding boxes of the objects in the generated images. The consistency of the detection results with the object-wise guidance information can represent the effectiveness of the object-wise controlling, specifically, given a generated image and the object-wise guidance $\{\mathcal{B}, \mathcal{C}\}$ it uses, for each entry $\{b, c\} \in \{\mathcal{B}, \mathcal{C}\}$, we first find out all detected objects with the class label $c$ and record the maximum IoU. If no object with the label $c$ is detected, IoU corresponding to $\{b,c\}$ is recorded as $0$. If the recorded IoU is more significant than $0.5$, $\{b, c\}$ is regarded as a successful control. 

We notice that the size and the position of the guidance box have a significant impact on the control efficiency. We categorize all objects into three subsets based on the size of the guidance bounding box. Specifically, the subset $S$ contains all objects with an area less than $150^2$, the subset $L$ contains all objects with an area greater than $300^2$, while the other objects are channeled into subset $M$. On the other hand, We evenly divide all guidance boxes into three subsets, $near$, $mid$, and $far$, based on the distance between the center of the guidance box and the center of the image and evaluate the control effectiveness of each subset.

\noindent\textbf{Image Quality} FID~\cite{heusel2017gans} is a measure of similarity between two datasets of images. We evaluate the quality of generated images by their FID with the ground truth images in datasets, and a smaller FID indicates a better image quality.

\vspace{-5pt}
\section{Results and Analysis}
\vspace{-5pt}
We first show overall results in \Tref{tab:results}. Our proposal outperforms the stable diffusion on object-wise consistency (IoU, $R_{suc}$) thanks to the designed soft-mask while maintaining or even improving the image quality (FID). 
However, Paint ~\cite{balaji2022ediffi} controls the position of the generated object at the cost of image quality. Moreover, our proposal ($w^{\prime} = 0.15$ and $w^{\prime} = 0.2$) outperforms Paint on all settings of $w^{\prime}$. 
We then show the evaluation results on the control effectiveness of different subsets considering guidance size and position in \Tref{tab:distance}, from which we can find that our proposal outperforms the related works on all conditions.

\noindent\textbf{Discussion on Guidance Size} The guidance size significantly affects the control's effectiveness. According to the results for each subset $s, m, l$, larger guidance bounding boxes will have larger IoUs and are more likely to be successful controlled. In contrast, the guidance of generating small objects is much more difficult.

\noindent\textbf{Discussion on Guidance Position} Although stable diffusion cannot accept object-wise guidance, generated objects sometimes overlap with the guidance box by chance when the object guidance box is relatively large and near to the center of image, because the model tends to generate objects in the center of the image. For objects far from the center, the likelihood of the stable diffusion accidentally generating objects at the specified location decreases significantly. In contrast, our proposed method maintains higher control over all subsets than the related works. 

\vspace{-5pt}
\section{Conclusions}
\vspace{-5pt}
This paper proposes an effective object generation position guiding method for stable diffusion and a method to evaluate object-wise generation consistency. We introduce a Gaussian distribution-based soft-mask to reduce the negative impact on the attention map while preserving the control efficiency. Experimental results show that our method achieves state-of-the-art performance on both control efficiency and image quality.

\bibliographystyle{IEEE}
\bibliography{refs}

\end{document}